%% file: camera_ready.tex
\documentclass[conference]{IEEEtran}
\IEEEoverridecommandlockouts
\pdfoutput=1
\usepackage{cite}
\usepackage{amsmath,amssymb,amsfonts}
\usepackage{graphicx}
\usepackage{textcomp}
\usepackage{xcolor}
\usepackage[
    a4paper,
    total={184mm,239mm}
]{geometry}

\def\BibTeX{%
    {\rm B\kern-.05em%
    {\sc i\kern-.025em b}%
    \kern-.08em T\kern-.1667em%
    \lower.7ex\hbox{E}%
    \kern-.125emX}%
}

\usepackage{booktabs} 
\usepackage{gensymb}
\usepackage[english]{babel}
\usepackage{bm}
\usepackage[capitalise]{cleveref}
\usepackage{subcaption}
\usepackage{pifont}
\usepackage{makecell}
\usepackage{multirow}
\usepackage{multicol}
\usepackage{algorithm}
\usepackage{algpseudocode}

\newcommand{\Vmin}{V_{min}}

\newcommand{\InputSpace}{\mathcal{X}}
\newcommand{\OutputSpace}{\mathcal{Y}}

\newcommand{\model}{g}

\newcommand{\x}{\bm{\mathrm{x}}}

\newcommand{\y}{\bm{\mathrm{y}}}

\newcommand{\h}{\bm{\mathrm{h}}}

\newcommand{\W}{\bm{{W}}}

\DeclareMathOperator*{\argmin}{arg\,min}

\makeatletter
\newcommand{\smallbullet}{} 
\DeclareRobustCommand\smallbullet{%
  \mathord{\mathpalette\smallbullet@{0.7}}%
}
\newcommand{\smallbullet@}[2]{%
  \vcenter{\hbox{\scalebox{#2}{$\m@th#1\bullet$}}}%
}
\makeatother

\begin{document}

\makeatletter
\newcommand{\linebreakand}{%
  \end{@IEEEauthorhalign}
  \hfill\mbox{}\par
  \mbox{}\hfill\begin{@IEEEauthorhalign}
}
\makeatother

\title{
  Transfer Learning for Minimum Operating Voltage Prediction in Advanced Technology Nodes: Leveraging Legacy Data and Silicon Odometer Sensing
}

\author{
\IEEEauthorblockN{Yuxuan Yin\IEEEauthorrefmark{2}, Rebecca Chen\IEEEauthorrefmark{3}, Boxun Xu\IEEEauthorrefmark{2}, Chen He\IEEEauthorrefmark{3}, Peng Li\IEEEauthorrefmark{2}\\
\IEEEauthorblockA{\IEEEauthorrefmark{2}Department of Electrical and Computer Engineering, University of California Santa Barbara, CA, USA}
\IEEEauthorblockA{\IEEEauthorrefmark{3}Automotive Processing, NXP Semiconductors, TX, USA \\}
\IEEEauthorblockA{\{y\_yin, boxunxu, lip\}@ucsb.edu, \{rebecca.chen\_1, chen.he\}@nxp.com}}
}
\maketitle

\begin{abstract}
Accurate prediction of chip performance is critical for ensuring energy efficiency and reliability in semiconductor manufacturing. However, developing minimum operating voltage ($V_{min}$) prediction models at advanced technology nodes is challenging due to limited training data and the complex relationship between process variations and $V_{min}$. To address these issues, we propose a novel transfer learning framework that leverages abundant legacy data from the 16nm technology node to enable accurate $V_{min}$ prediction at the advanced 5nm node. A key innovation of our approach is the integration of input features derived from on-chip silicon odometer sensor data, which provide fine-grained characterization of localized process variations—an essential factor at the 5nm node—resulting in significantly improved prediction accuracy.
\end{abstract}

\section{Introduction}

The minimum operating voltage ($V_{\text{min}}$) constitutes a critical parameter that defines the lowest voltage threshold at which an integrated circuit can maintain reliable functionality while meeting specified performance criteria. This fundamental characteristic plays a pivotal role in determining both the power efficiency and operational stability of semiconductor devices, with implications spanning a broad spectrum of applications from portable and automotive electronics to large-scale data centers. Furthermore, structural test methodologies for $V_{\text{min}}$ characterization, including SCAN patterns and SRAM Built-In Self Test (MBIST), have emerged as increasingly sophisticated and crucial tools for identifying subtle manufacturing defects, particularly in advanced technology nodes where process variations become more pronounced \cite{VminTest}.

However, measuring accurate $V_{\text{min}}$ is challenging due to the high cost and time required for the test, which is unacceptably expensive for production testing. As a result, semiconductor manufacturers rely on surrogate metrics to train a model for $V_{\text{min}}$ prediction. In legacy technology nodes, such as 16nm and below, typical surrogate metrics include the Process Observation Structure (POSt) and parametric test data, and predicting models are often built leveraging various machine learning techniques such as random forests, Gaussian process or gradient-tree approaches like XGBoost \cite{XGBoost} and CatBoost \cite{CatBoost}, as well as deep learning models, including multilayer perceptions (MLP) and monotonic lattice networks \cite{yin-mln-itc,deepLattice,LatticeRegression}. As technology nodes advance to 5nm and below, two new challenges arise: 1) process variations become more pronounced, and 2) available training data is limited. These challenges make the correlations between surrogate metrics and $V_{\text{min}}$ weaker and the prediction models nongeneralizable to new technology nodes, resulting in poor prediction performance.

Firstly, the semiconductor industry's continuous drive towards miniaturization has led to increasingly significant process variations in advanced technology nodes. These variations manifest in multiple forms, including random dopant fluctuations (RDF), line edge roughness (LER), and gate length variability, which become more pronounced at smaller scales. For instance, in 5nm technology nodes, even minor deviations in the manufacturing process can result in substantial variations in transistor characteristics, affecting critical parameters such as threshold voltage and leakage current. This variability is further exacerbated by the three-dimensional nature of modern FinFET and Gate-All-Around (GAA) architectures, where process control becomes more challenging. The impact of these variations is particularly evident in the context of $V_{\text{min}}$ prediction, where the complex interplay between different variation sources can lead to significant deviations in chip performance and reliability characteristics.

Secondly, the scarcity of data in advanced technology nodes presents a significant challenge for developing accurate $V_{\text{min}}$ prediction models. This limitation stems from several factors: the high cost of chip manufacturing in advanced nodes, which restricts the number of test chips available; the extended development cycles required for new processes; and the time-intensive nature of comprehensive testing procedures. For example, while a mature 16nm node might have thousands of characterized chips available for model training, a newly developed 5nm node might only have data from a few hundred chips. This order-of-magnitude reduction in training data significantly impacts the ability to train robust machine learning models, particularly deep learning approaches that typically require substantial amounts of data to achieve good generalization. Furthermore, the data scarcity issue is exacerbated by the increased complexity of advanced nodes, where more sophisticated models are needed to capture the intricate relationships between process variations and chip performance, yet insufficient data is available to train such models effectively.

\begin{figure}[t]
    \centering
    \includegraphics[width=\linewidth]{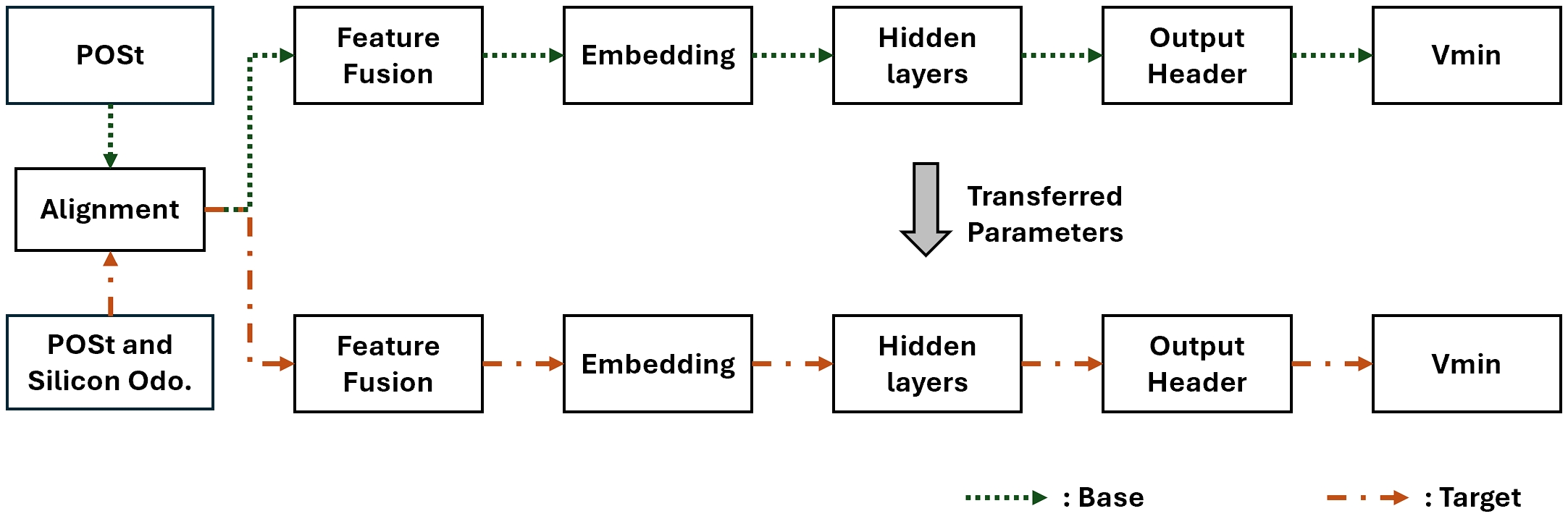}
    \caption{The workflow of the proposed transfer learning $\Vmin$ prediction framework}
    \label{fig:flow-chart}
\end{figure}

To address these challenges, we propose a novel transfer learning framework that leverages both legacy data from mature technology nodes and advanced on-chip silicon odometer sensing capabilities. As illustrated in \cref{fig:flow-chart}, our approach utilizes the abundant data available from 16nm technology to establish a robust foundation model, which is then adapted to the 5nm node through careful fine-tuning. We leverage domain knowledge from our engineering expertise to group features based on device and function types, aligning them effectively across technology nodes through learnable linear transformations. Additionally, we incorporate silicon odometer sensors, which are specifically designed to measure process variations in advanced nodes, providing more direct correlations to chip performance compared to traditional surrogate metrics. 

This combination of transfer learning and enhanced sensing capabilities enables us to overcome both the data scarcity limitation and the magnified process variation challenges inherent in $V_{\text{min}}$ prediction of advanced technology nodes. To validate our proposed framework, extensive experimental analyses were performed on industrial chip datasets from both 16nm and 5nm technologies. The results demonstrate that transfer learning significantly improves prediction accuracy, especially when leveraging silicon odometer features compared to traditional process-related features.

The key contributions of this work are:
\begin{itemize}
    \item We propose a novel transfer learning framework that leverages legacy data from mature 16nm technology nodes to enhance $V_{\text{min}}$ prediction in advanced 5nm nodes, effectively addressing the data scarcity challenge while maintaining high prediction accuracy.
    \item We integrate silicon odometer sensors specifically designed for advanced nodes, providing more direct correlations to chip performance compared to traditional surrogate metrics and enabling better characterization of process variations.
    \item We demonstrate through extensive experiments on industrial datasets that our approach significantly improves prediction accuracy, particularly when combining transfer learning with silicon odometer features, establishing a scalable solution for future semiconductor technologies.
\end{itemize}

\section{Related Work}
\input{sec/related_work}

\section{Preliminaries}
\input{sec/prelim}

\section{Methodology}
\input{sec/method}

\section{Experimental Results}
\input{sec/expr}

\section{Conclusion}
In this work, we present a novel transfer learning framework for predicting the minimum operating voltage, \(V_{\min}\), of automotive chips. Our approach bridges the gap between mature 16nm and advanced 5nm nodes by leveraging extensive legacy data alongside innovative silicon odometer sensing. We make three key contributions: integrating legacy data through neural network pretraining and fine-tuning, enhancing prediction accuracy with silicon odometer features that capture subtle process variations, and demonstrating strong scalability across varying data conditions and technology nodes.

Our framework provides a robust and efficient solution for \(V_{\min}\) prediction by effectively combining legacy data and advanced sensor inputs, with promising experimental results that validate the potential of transfer learning in advanced semiconductor technologies. Future research will explore enhancements in domain adaptation techniques and integration of additional sensor modalities to further advance semiconductor performance modeling.

\section*{Acknowledgment}
This work is supported by an NXP Long Term University (LTU) grant and the National Science Foundation under Grant No. 1956313 and No. 2334380.

\bibliographystyle{IEEEtran}
\bibliography{IEEEabrv, ref}

\end{document}

%% file: sec/related_work.tex
\subsection{Surrogate Metrics for \(V_{\min}\) Correlation}
Measuring \(V_{\min}\) directly through functional testing is time-consuming and expensive, often requiring specialized equipment and test patterns. To address this challenge, some low-cost surrogate metrics have been leveraged to predict \(V_{\min}\) values. Typical surrogate metrics include Process Observation Structures (POSt) and parametric test data. POSts are dedicated test structures embedded in the chip that measure key electrical parameters like transistor threshold voltage, gate delay, and leakage current. Parametric test data includes measurements of critical path delays, power consumption, and other electrical characteristics obtained during wafer-level testing. Both POSt measurements and parametric test data can be collected quickly and cost-effectively, such that they are employed to model chip performance for 16nm and below in current semiconductor manufacturing practice.

For more advanced technology nodes, the process variation is more severe, making the relationship between \(V_{\min}\) and the surrogate metrics more complex. Therefore, it is insufficient to rely exclusively on conventional surrogate metrics such as POSt measurements and parametric test data to predict \(V_{\min}\) values. Silicon odometer sensors, i.e., on-chip monitors such as ring oscillators and delay chains, are invented and deployed in recent years to provide better correlation with \(V_{\min}\) and $F_{\max}$ in real-time \cite{SiOdo2008,SiOdo2010, SiOdo2011}. While these monitors are designed for voltage and frequency degradation measurement \cite{yin-mln-itc,yin-date2024}, their capability of predicting the chip performance at the manufacturing stage has not been well studied. To this end, we are motivated to analyze the relationship between silicon odometers and \(V_{\min}\), in addition to the existing surrogate metrics including POSts and parametric test data.

\subsection{Machine Learning Approaches in Chip Performance \\ Modeling}
Traditional statistical approaches have been widely used for \(V_{\min}\) modeling in semiconductor manufacturing. Linear regression models were among the first methods applied to this problem, establishing baseline correlations between process parameters and \(V_{\min}\) \cite{yin-vts-2024-rba,functionalATE}. While these linear models can make predictions with little computational overhead, which is critical in the manufacturing stage, their simplicity can be a limitation when dealing with complex nonlinear relationships that emerge in advanced technology nodes.

Apart from linear regression, nonlinear models have been explored. Gaussian Process (GP) regression \cite{GP} has been adopted in recent years as a more sophisticated statistical approach that can capture nonlinear relationships while providing uncertainty estimates \cite{GPModel-Fmax}. Ensemble methods, such as Random Forests and Gradient Boosting Machines like XGBoost \cite{XGBoost} and CatBoost \cite{CatBoost}, have also proven effective for \(V_{\min}\) degradation prediction in \cite{yin-date2024, yin-dac2024}. These approaches combine multiple weak learners to create a robust predictor that can handle the inherent noise and randomness caused by process variation in semiconductor manufacturing. However, both GP and ensemble methods can degrade in limited data settings due to the curse of dimensionality, and they are hard to integrate with transfer learning to improve the prediction accuracy by leveraging the abundant data from legacy technology nodes.

More recently, artificial neural networks have been applied to \(V_{\min}\) prediction tasks \cite{yin-date2024, yin-mln-itc}. For instance, \cite{yin-mln-itc} adopted a Monotonic Lattice Network (MLN) to predict \(V_{\min}\) in addressing the nature of monotonicity constraint of \(V_{\min}\) degradation. Vanilla Multi-Layer Perceptron (MLP) models have also been applied to \(V_{\min}\) prediction tasks \cite{yin-date2024,yin-mln-itc} as a baseline model, but their performance is not competitive due to the limited data, overfitting issue, and the design of the network architecture. To address these challenges, we propose a transfer learning-based approach to improve the prediction accuracy by leveraging the abundant data from legacy technology nodes. We also adopt normalization techniques to improve the robustness and generalization ability of the model.

%% file: sec/prelim.tex
\subsection{\(V_{\min}\) Prediction}
\noindent The minimum operating voltage, \(V_{\min}\), is a critical parameter that defines the lowest supply voltage at which a chip can operate reliably. Accurate prediction of \(V_{\min}\) is vital for optimizing power management and ensuring chip performance under various process variations, especially in automotive applications where safety and reliability are paramount.

Given a set of measured features \(\mathbf{x} \in \mathbb{R}^d\) (e.g., Process Observation Structures (POSt) features and parametric test data) and a certain $V_{min}$ pattern evaluation $y \in (0, +\infty)$ (e.g., DC Scan, AC Scan, MBIST $V_{min}$), the task of \(V_{\min}\) prediction can be formulated as a regression problem:

\begin{equation}
    y = g(\mathbf{x}; \bm{\theta}) + \epsilon,
\end{equation}

where \(g(\mathbf{x}; \bm{\theta})\) is a predictive model parameterized by \(\bm{\theta}\). \(\epsilon\) denotes the error term, typically assumed to be a small white noise. 

The goal is to learn the mapping \(g\) such that the prediction \(\hat{y} := g(\mathbf{x}; \bm{\theta})\) is as close as possible to the true \(y\). To achieve this, the model parameters \(\theta\) are estimated by minimizing a loss function, commonly the Mean Squared Error (MSE):

\begin{equation}\label{eq:mse-loss}
    \mathcal{L}(\bm{\theta}; \{\mathbf{x}_i, \y_i\}_{i=1}^N) = \frac{1}{N} \sum_{i=1}^{N} \left(y_{i} - g(\mathbf{x}_i; \theta)\right)^2,
\end{equation}

where \(N\) is the number of training samples and \(y_{i}\) is the measured minimum operating voltage for the \(i\)-th chip. 

While traditional linear regression models assume a linear relationship between \(\mathbf{x}\) and \(V_{\min}\), they often fail to capture the complex, nonlinear interactions caused by process variations, particularly in advanced technology nodes.
\begin{figure}[htbp]
    \centering
    \includegraphics[width=\linewidth]{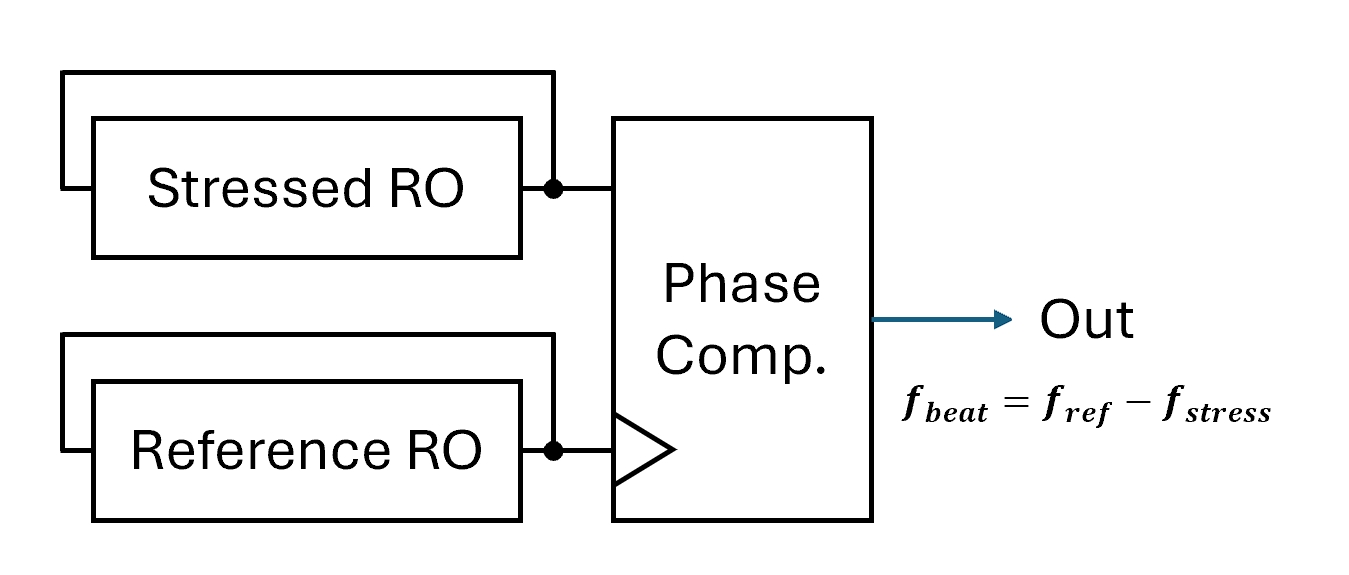}
    \caption{High level diagram of a silicon odometer \cite{SiOdo2011}.}
    \label{fig:SiOdo}
\end{figure}
\subsection{Silicon Odometers}

Silicon odometers \cite{SiOdo2008,SiOdo2010,SiOdo2011} are on-chip monitoring structure consisting of a pair of ring oscillators (ROs): a stressed RO and a reference RO. \cref{fig:SiOdo} shows the high level diagram of a silicon odometer: the reference RO serves as a baseline, while changes in the stressed RO's frequency relative to this baseline indicate aging effects. The frequency difference between these two oscillators, known as the beat frequency \(f_{beat}\), provides a measure of circuit aging and performance degradation. This differential measurement approach helps filter out common-mode noise and environmental variations, making silicon odometers particularly effective for real-time monitoring of circuit degradation.

\begin{figure}[htbp]
    \center\includegraphics[width=\linewidth]{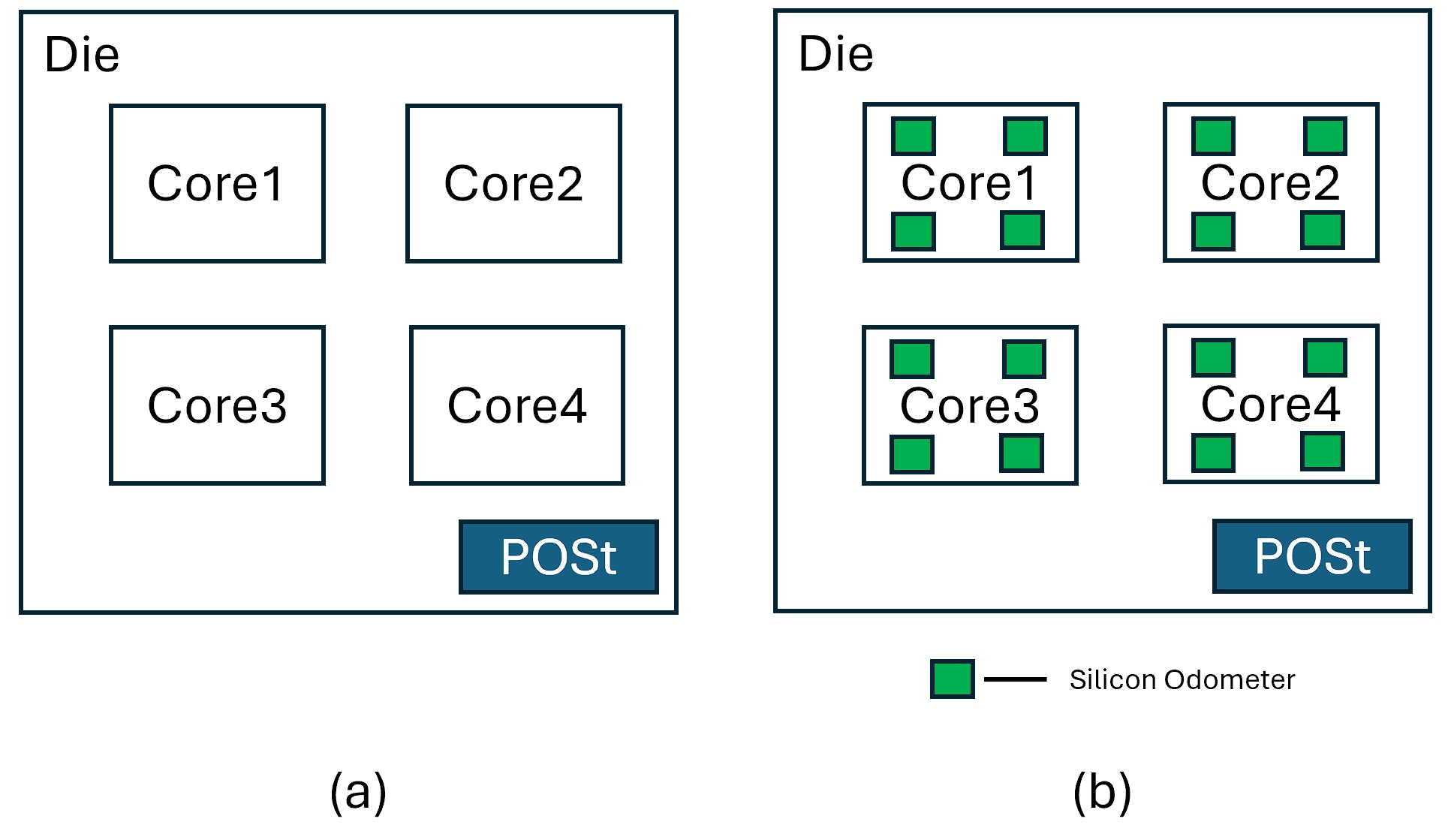}
    \caption{Illustration of POSt and silicon odometer placement: (a) 16nm multi-core product with only POSt, and (b) 5nm multi-core product with both POSt and silicon odometers. }
    \label{fig:POSt_SiOdo}
\end{figure}

\begin{figure*}[t]
    \centering
    \includegraphics[width=\linewidth]{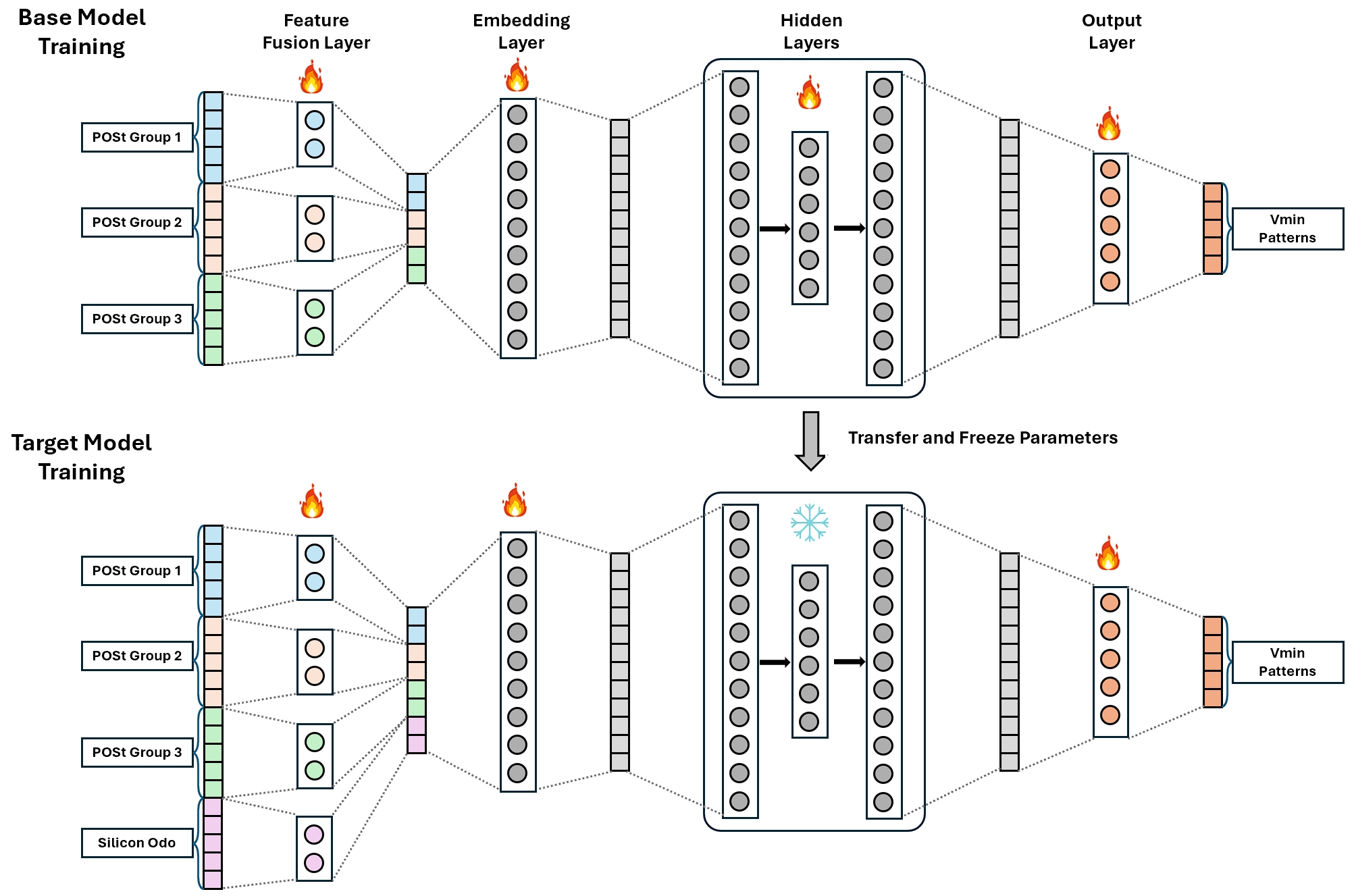}
    \caption{Proposed transfer learning framework for $V_{\text{min}}$ prediction. The square nodes represent the features, and the circle nodes represent the compute units of the model.}
    \label{fig:framework}
\end{figure*}

A key advantage of silicon odometers lies in their ability to provide fine-grained, localized performance monitoring across the chip. As illustrated in \cref{fig:POSt_SiOdo}, while traditional POSt structures are typically limited to a single set of test structures at the corner of each SOC/die providing only global process feedback, silicon odometers can be distributed throughout each core. This distributed monitoring approach enables more effective capturing of within-die characteristics and spatial variations in circuit performance.

In our task, which is initial (T0) $V_{\min}$ estimation rather than tracking aging effects, we propose to leverage silicon odometers as input features to better capture localized process variations.  At T0, before any aging has occurred, the stressed and reference ring oscillators in each silicon odometer produce identical frequencies, allowing either to be used as a process monitor for $V_{\min}$ prediction. 

\subsection{Transfer Learning}
\noindent Transfer learning is a machine learning approach that leverages knowledge from a source domain with abundant labeled data to improve performance on a target domain with limited data. In our context, the source domain consists of a mature technology node (e.g., a 16nm node) and the target domain of an advanced node (e.g., a 5nm node). Let us denote the source domain dataset as
\[
\mathcal{D}_S = \{ (\mathbf{x}_S^{(i)}, \mathrm{y}_S^{(i)}) \}_{i=1}^{N_S},
\]
and the target domain dataset as
\[
\mathcal{D}_T = \{ (\mathbf{x}_T^{(j)}, \mathrm{y}_T^{(j)}) \}_{j=1}^{N_T},
\]
with \(N_S \gg N_T\). The primary aim is to learn a predictive function that generalizes well on \(\mathcal{D}_T\) by transferring useful representations learned from \(\mathcal{D}_S\).

A common transfer learning approach is to pretrain a neural network on \(\mathcal{D}_S\) and then fine-tune the model using the limited target domain data. Formally, the pretrained parameters \(\bm{\theta}_S\) are obtained by minimizing the loss on the source domain:
\begin{equation}
    \bm{\theta}_S = \arg\min_{\theta} \frac{1}{N_S} \sum_{i=1}^{N_S} \ell\left(\mathrm{y}_S^{(i)}, f(\mathbf{x}_S^{(i)}; \theta)\right),
\end{equation}
where \(\ell(\cdot,\cdot)\) is typically the MSE loss for regression tasks and the cross-entropy loss for classification tasks. These parameters are used to initialize the target model parameters \(\bm{\theta}_T\) for the target domain. Furthermore, to mitigate the impact of domain shifts between \(\bm{\theta}_S\) and \(\mathcal{D}_T\), a domain adaptation term can be incorporated into the training objective:
\begin{equation}
    \mathcal{L}_{\text{total}} = \mathcal{L}_T(\bm{\theta}_T;\mathcal{D}_T) + \lambda \cdot \mathcal{R}(\bm{\theta}_T, \bm{\theta}_S),
\end{equation}
where \(\mathcal{L}_T(\theta)\) is the target domain loss, \(\mathcal{R}(\cdot,\cdot)\) regularizes the discrepancy between the source and target models, and \(\lambda\) is a regularization parameter balancing the two terms.

In the application of \(V_{\min}\) prediction, transfer learning offers significant benefits by enabling the model to leverage extensive data from older technology nodes. This not only improves prediction accuracy but also enhances data efficiency and generalization in advanced nodes with limited training samples.

%% file: sec/method.tex
Our proposed transfer learning framework for \(V_{\min}\) prediction consists of three main components: (1) feature engineering and alignment across technology nodes, (2) model architecture design for transfer learning, and (3) training strategy optimization. The framework leverages both abundant legacy data and silicon odometer sensors to improve prediction accuracy. A detailed illustration of the framework is shown in \cref{fig:framework}. Without loss of generality, the base technology node is 16nm and the target technology node is 5nm in this paper, and the products considered in both nodes are automotive chips.

\subsection{Problem Formulation}
Let $\InputSpace^{(b)} \in \mathbb{R}^{m^{(b)}}$ and $\OutputSpace^{(b)} \in \mathbb{R}^{o^{(b)}}$ denote the input feature space and output space respectively for the base technology node. The input space is the set of $m^{(b)}$ POSt and parametric test features and the output space is the set of $o^{(b)}$ $\Vmin$ test patterns, such as DC Scan, AC Scan, and M-BIST $\Vmin$ with different testing conditions. Similarly, let $\InputSpace^{(t)} \in \mathbb{R}^{m^{(t)}}$ and $\OutputSpace^{(t)} \in \mathbb{R}^{o^{(t)}}$ denote the input feature space of $m^{(t)}$ POSt, parametric test and silicon odometer features and the output space of $o^{(t)}$ $\Vmin$ test patterns for the target technology node, respectively.

Suppose that we have a large dataset $\mathcal{D}^{(b)} = \{(\x_i^{(b)}, \y_i^{(b)})\}_{i=1}^{n^{(b)}}$ of $n^{(b)}$ samples for the base technology node and a limited dataset $\mathcal{D}^{(t)} = \{(\x_i^{(t)}, \y_i^{(t)})\}_{i=1}^{n^{(t)}}$ of $n^{(t)} \ll n^{(b)}$ samples for the target technology node. Given a base model $\model^{(b)}(\cdot; \bm{\theta}^{(b)}): \InputSpace^{(b)} \to \OutputSpace^{(b)}$ trained on the base dataset, where $\bm{\theta}^{(b)}$ is the set of parameters of the base model, our goal is to learn a new model $\model^{(t)}(\cdot; \bm{\theta}^{(t)}): \InputSpace^{(t)} \to \OutputSpace^{(t)}$ that minimizes the prediction error on the target dataset while leveraging knowledge from the base model. The model parameters $\bm{\theta}^{(t)}$ are optimized through:

\begin{equation}\label{eq:problem_formulation}
    \bm{\theta}^{(t)} = \argmin_{\bm{\theta}^{(t)}} \frac{1}{n^{(t)}} \sum_{i=1}^{n^{(t)}} \left(\y_i^{(t)} - \model^{(t)}(\x_i^{(t)}; \bm{\theta}^{(t)})\right)^2 + \mathcal{R}(\bm{\theta}^{(t)}, \bm{\theta}^{(b)})
\end{equation}
where the first term is the MSE loss defined in (\ref{eq:mse-loss}) that quantifies the $\Vmin$ prediction error, and the second one is a regularization term that encourages the target model parameters $\bm{\theta}^{(t)}$ to be close to the parameters $\bm{\theta}^{(b)}$ of the base model.

\subsection{Feature Engineering and Alignment}
The feature sets between 16nm and 5nm technology nodes differ significantly due to process evolution and the introduction of new measurement capabilities. We categorize the features into three groups, and treat them differently in the transfer learning process:

\begin{itemize}
    \item \textbf{Common Features}: Surrogate features that exist in both nodes, such as threshold voltage and leakage current measurements from POSts. After normalization, these features undergo a grouping process to their designed functional types, which are manually defined by domain experts. Each group is mapped to a number of $k$ hidden features in both the base and target networks. More details about the feature fusion is presented in \cref{sec:model_architecture}.
    
    \item \textbf{Legacy Features}: Parameters specific to the 16nm technology that have no direct correspondence in 5nm. These features are not transferable so that they are abandoned from the training of the base neural network.
    
    \item \textbf{Silicon Odometer Features}: New silicon odometer measurements available only in 5nm technology. These features provide enhanced correlation with \(V_{\min}\) in advanced nodes. Like the common features, all the silicon odometer sensor data are treated as a single group so that they are fused into $k$ hidden features in the target network.
\end{itemize}

Another challenge in training a $\Vmin$ prediction model is the difference in feature scales across testing structures and sensors. For instance, the ring oscillator frequency has a unit in GHz while the leakage current has a unit in nA. This large-scale difference can lead to unstable training and poor generalization, especially in our transfer learning scenarios. To align features across nodes and stabilize the transfer learning process, we further employ normalization in either technology node. For instance, given an input sample $\x^{(b)} \in \mathbb{R}^{m^{(b)}}$ for the base technology node, the normalized feature $\hat{\x}^{(b)} \in \mathbb{R}^{m^{(b)}}$ is computed as:

\begin{equation}\label{eq:norm}
    \hat{\x}^{(b)}[j] = \frac{\x^{(b)}[j] - \min(\{\x^{(b)}_{i}[j]\}_{i=1}^{m^{(b)}})}{\max(\{\x^{(b)}_{i}[j]\}_{i=1}^{m^{(b)}}) - \min(\{\x^{(b)}_{i}[j]\}_{i=1}^{m^{(b)}})}
\end{equation}

where $[\cdot]$ is the index of the feature, and $\max\{\x^{(b)}_{i}[j]\}_{i=1}^{m^{(b)}}$ and $\min\{\x^{(b)}_{i}[j]\}_{i=1}^{m^{(b)}}$ are the maximum and minimum values of the $j$-th feature across all samples in the base dataset.

\subsection{Model Architecture} \label{sec:model_architecture}
Our base and target model are both artificial neural networks with multiple fully-connected layers. They share the same structure: 

\begin{itemize}
    \item \textbf{Feature Fusion Layer $l_{fusion}(\cdot; \bm{\theta}_{fusion})$}: Extracts each functional group of input features into a number of $k$ hidden features.
    \item \textbf{Embedding Layer $l_{embedding}(\cdot; \bm{\theta}_{embedding})$}: Embeds the fused features into an $m^{emb}$-dimensional embedding space.
    \item \textbf{Hidden Layers $l_{hidden}(\cdot; \bm{\theta}_{hidden})$}: Contains multiple fully-connected layers with Leaky ReLU activation functions that process the fused features. These layers learn hierarchical representations of the input features and enable knowledge transfer between technology nodes.
    \item \textbf{Output Layer $l_{output}(\cdot; \bm{\theta}_{output})$}: Generates the final \(V_{\min}\) prediction based on the processed features. It is a fully-connected layer with $o^{(b)}$ and $o^{(t)}$ output neurons for the base and target networks respectively.
\end{itemize}

\subsubsection{Feature Fusion Layer}
The feature fusion layer $l_{fusion}(\cdot; \bm{\theta}_{fusion})$ maps each functional group of features into a fixed number of $k$ hidden features through learnable linear transformations. For either technology node, suppose we have $g$ functional groups of features, and the $i$-th group contains $m_i$ features. The fusion layer transforms each group into $k$ hidden features through:

\begin{equation}
    \h_i = \W_i \x_i + \bm{b}_i
\end{equation}
where $\x_i \in \mathbb{R}^{m_i}$ is the input features of the $i$-th group, $\W_i \in \mathbb{R}^{k \times m_i}$ and $\bm{b}_i \in \mathbb{R}^k$ are the learnable weight matrix and bias vector, respectively. The fused features from all groups are concatenated to form the final output of the fusion layer:

\begin{equation}
    l_{fusion}(\x; \bm{\theta}_{fusion}) = [\h_1; \h_2; \cdots; \h_{g}] \in \mathbb{R}^{k \times g}
\end{equation}

Our feature fusion layer is designed to compress the information of each functional group into a fixed number of $k$ hidden features, which is a trade-off between the model complexity and the information loss. Compared to previous works \cite{yin-date2024,yin-mln-itc,yin-dac2024} where a feature selection process is applied to select the most important features, which is model-agnostic, our feature fusion layer is optimized through the model training process to best extract the information of each functional group.

\subsubsection{Embedding Layer}
The dimension of the fused features is different for different technology nodes, due to the introduction of silicon odometer features to the advanced technology node. To better facilitate transfer learning, we map the fused features of different technology nodes into a unified $m^{emb}$-dimensional embedding space, which is a fully-connected layer with a Leaky ReLU activation function.

\subsubsection{Hidden Layers}
The hidden layers $l_{hidden}(\cdot; \bm{\theta}_{hidden})$ aims to learn transferable representations of the embedding features from different technology nodes. It is designed as a series of fully-connected layers with Leaky ReLU activation functions.

\subsubsection{Output Layer}
The output layer $l_{output}(\cdot; \bm{\theta}_{output})$ generates the final \(V_{\min}\) prediction based on the final representations of input features. It is a fully-connected layer with $o^{(b)}$ and $o^{(t)}$ output neurons for the base and target networks, respectively.

\subsection{Training Strategy}
In our transfer learning framework, we first train the base model on the 16nm dataset and then fine-tune the target model on the 5nm dataset. 

\subsubsection{Base Model Training}
The base model is trained on the 16nm dataset using the following loss function:

\begin{equation}
    \mathcal{L}_{mse}(\bm{\theta}^{(b)};\mathcal{D}^{(b)}) = \frac{1}{n^{(b)}\cdot o^{(b)}} \sum_{i=1}^{n^{(b)}} \|\y_i^{(b)} - \model^{(b)}(\x_i^{(b)}; \bm{\theta}^{(b)})\|_2^2
\end{equation}
where $\y_i^{(b)} \in \mathbb{R}^{o^{(b)}}$ is the ground truth \(\Vmin\) test patterns and $\x_i^{(b)} \in \mathbb{R}^{m^{(b)}}$ is the input features. We average the loss across all test patterns and samples in the 16nm dataset to train our base model. This setting is different from previous works \cite{yin-date2024,yin-mln-itc,yin-dac2024} where each $\Vmin$ pattern has its single-output predictive model. Our model is able to learn better representations of the input features and prevent overfitting problems on target domain datasets with limited volume.

\subsubsection{Target Model Training}
Our target model leverages the learned parameters from the base model and fine-tunes on the 5nm dataset. We set the parameters of the hidden layers of the target model $\bm{\theta}_{hidden}^{(t)}$ to be the same as the base model $\bm{\theta}_{hidden}^{(b)}$, and freeze them during the training process. The optimization problem is formulated as:


\begin{equation}
    \bm{\theta}^{(t)} \setminus \bm{\theta}_{hidden}^{(t)} = \argmin_{\bm{\theta}^{(t)} \setminus \bm{\theta}_{hidden}^{(t)}} \mathcal{L}_{mse}(\bm{\theta}^{(t)};\mathcal{D}^{(t)})
\end{equation}

The solution of this optimization problem is equivalent to the one in (\ref{eq:problem_formulation}), when the regularization term $\mathcal{R}(\bm{\theta}^{(t)}, \bm{\theta}^{(b)})$ is defined as:

\begin{equation}
    \mathcal{R}(\bm{\theta}^{(t)}, \bm{\theta}^{(b)}) = \begin{cases}
        0 & \text{if } \bm{\theta}_{hidden}^{(t)} = \bm{\theta}_{hidden}^{(b)} \\
        \infty & \text{otherwise}
    \end{cases}
\end{equation}

As the learnable layers of the target model consist of feature fusion, embedding, and output layers, which are all shallow linear transformations, they can be optimized on a small target dataset without overfitting.

%% file: sec/expr.tex
\subsection{Experimental Settings}
We conduct comprehensive experiments to evaluate our proposed transfer learning framework for $\Vmin$ prediction on a large dataset of 16nm automotive chips and a limited dataset of 5nm automotive chips.

\begin{table}[htbp]
    \centering
    \caption{Description of 16nm and 5 nm industrial datasets}
    \begin{tabular}{l |c c }
    \toprule
        Attributes & 16nm& 5nm\\
    \midrule
        \# Dies & 5239& 415\\
        \# Grouped POSt Features& 5, 19, 21& 12, 7, 18\\
 \# Silicon Odometers& -&124\\
        \# $\Vmin$ patterns&  63& 27  \\
        Test temperature / \degree C & -40, 25, 125 & -45, 25, 80, 125\\
    \bottomrule
    \end{tabular}
    \label{tab:dst-16}
\end{table}

\textbf{16 nm dataset:} As shown in \cref{tab:dst-16}, our 16 nm dataset has a total number of 5,239 measured chips.  For each chip, we employ the same test flow to collect POSt features and $\Vmin$, used for input data and output targets of the prediction model, respectively. We group POSt features into 3 functional groups, whose numbers of features are 5, 19, and 21, respectively. In the testing flow, we first test all patterns of $\Vmin$, and then perform the parametric tests. Both phases are done at the same specific temperature: -45\degree C (cold), 25\degree C (room), or 125\degree C (hot). Both $\Vmin$ and parametric tests are performed on an Automatic Test Equipment (ATE) tester. A total of 63 patterns of $\Vmin$ are measured, including DC Scan, AC Scan, and MBIST $\Vmin$.

\textbf{5 nm dataset:} Our experiments use a limited dataset of 415 5nm automotive chips to demonstrate the effectiveness of the proposed $\Vmin$ prediction framework based on transfer learning. The test flow is identical to that used in the 16 nm technology node with slight variations in test settings. As outlined in \cref{tab:dst-16}, each chip undergoes measurement for 27 distinct patterns of $\Vmin$, covering DC Scan, AC Scan, and MBIST categories. Additionally, we utilize a richer set of parametric features for the 5 nm dataset, including 37 grouped POSt features and 124 silicon odometers, to enhance the sensitivity of the prediction model to subtle process variations at this advanced node. 27 $\Vmin$ patterns are measured, including DC Scan, AC Scan, and MBIST $\Vmin$. Tests are conducted across a wider temperature range—specifically, at -45\degree C (cold), 25\degree C (room), 80\degree C (warm), and 125\degree C (hot)— which is slightly different from the configurations used in the 16 nm dataset. The testing is executed on an ATE as well.

We describe our baseline methods as follows, and then present the configuration of our proposed transfer learning framework.

\textbf{Linear regression, XGBoost, and CatBoost:} Following \cite{yin-date2024,yin-mln-itc,yin-dac2024}, we apply Correlation Feature Selection (CFS) \cite{CFS} to select the three most relevant features for a certain $\Vmin$ pattern, and then use the selected features to train a baseline model. For XGBoost and CatBoost, we use the default parameters provided by the Python libraries of \cite{XGBoost,CatBoost}.

\textbf{Neural network:} For the neural network models, we use the Adam optimizer \cite{kingma2014adam} with a learning rate of $1 \times 10^{-3}$ and batch size of 32. The training process continues for 100k epochs. The proposed normalization method in \cref{eq:norm} is applied to all input features before the feature fusion layer. The feature fusion layer compresses each functional group into $k=2$ hidden features. The embedding layer maps the fused features into a 32-dimensional vector. The hidden layers consist of three fully-connected layers with 64, 16, and 64 neurons, respectively, all using Leaky ReLU activation functions \cite{leakyReLU}. The output layer generates predictions for all $\Vmin$ patterns simultaneously.

For our transfer learning process, we first train the base model on the 16nm dataset until convergence. Then, we copy the hidden layers of the base model to the target model and freeze them. We only fine-tune the feature fusion, embedding, and output layers on the 5nm dataset. The training process for the target model follows the same hyperparameters as the base model training for a fair comparison.

\subsection{Base Model Pretraining Results}
We first evaluate the performance of different models on the 16nm dataset. For all models, we randomly split the dataset into training (75\%) and testing (25\%) sets. To ensure fair comparison, we use the same data split for all models. The prediction target is the average $\Vmin$ across all patterns and temperatures for each die.
\begin{table}[htbp]
\centering
\caption{Prediction Results on 16nm Dataset}
\begin{tabular}{lccc}
\toprule
Tech & Model & Feature Type & Test RMSE (mV) \\
\midrule
16nm & Linear & POSt & 14.50 \\
16nm & XGBoost & POSt & 13.18 \\
16nm & CatBoost & POSt & 13.01 \\
16nm & Neural Network & POSt & \textbf{7.24} \\
\bottomrule
\end{tabular}
\label{tab:results-16nm}
\end{table}

As shown in \cref{tab:results-16nm}, our neural network model achieves the best performance with a test RMSE of 7.24mV, significantly outperforming baseline machine learning methods. The linear regression baseline yields a test RMSE of 14.50mV, while XGBoost and CatBoost achieve slightly better results with 13.18mV and 13.01mV, respectively. This demonstrates that the complex non-linear relationships between the POSt features and $\Vmin$ can be better captured by deep neural networks compared to conventional approaches. 

The superior performance of our neural network architecture can be attributed to several key design choices. First, the feature fusion layer effectively extracts information from different functional groups of the POSt features into a compact representation. Second, the embedding layer transforms the fused features into a rich, high-dimensional representation that captures subtle patterns in the data. Finally, the multi-layer architecture with Leaky ReLU activation enables modeling of complex non-linear relationships between the features and $\Vmin$ values.

These results on the 16nm dataset establish a strong foundation for our transfer learning approach, as the neural network has learned meaningful feature representations that may generalize to the 5nm technology node.


\subsection{Effectiveness of Transfer Learning and Silicon Odometer Features}
To evaluate the effectiveness of transfer learning and silicon odometer features in data-constrained scenarios, we conducted experiments using only 25\% of the 5nm dataset for training while maintaining the same test set. This setting simulates early development stages of a new technology node where labeled data is scarce.

\begin{table}[h!]
\centering
\caption{Prediction Results on 5nm Dataset}
\begin{tabular}{lccc}
\toprule
Tech & Model & Feature Type & Test RMSE (mV) \\
\midrule
5nm & Linear Model & POSt + siliconOdo & 7.09 \\
5nm & XGBoost & POSt + siliconOdo & 5.39 \\
5nm & CatBoost & POSt + siliconOdo & 4.59 \\
5nm & Neural Network & POSt + siliconOdo & 5.81 \\

5nm & Transferred NN & POSt + siliconOdo & \textbf{3.89} \\
\bottomrule
\end{tabular}
\label{tab:results-5nm-25}
\end{table}

As shown in \cref{tab:results-5nm-25}, the linear regression baseline achieves an $\Vmin$ prediction error of 7.09mV when using both the POSt and silicon odometer features. Traditional machine learning approaches like XGBoost and CatBoost show improved performance with $\Vmin$ prediction errors of 5.39 and 4.59mV, respectively. The neural network trained from scratch on this limited dataset achieves an $\Vmin$ prediction error of 5.81mV which is slightly better than the linear regression, but worse than XGBoost and CatBoost. This degradation highlights the challenge of training deep models from scratch with limited data from the target technology node.

However, our transfer learning approach demonstrates robust performance even with limited training data, which achieves the best performance with an RMSE of 3.89mV. This represents a significant improvement over the linear regression baseline and outperforms both traditional machine learning methods and the neural network trained from scratch.

\begin{figure}[htbp]
    \centering
    \includegraphics[width=\linewidth]{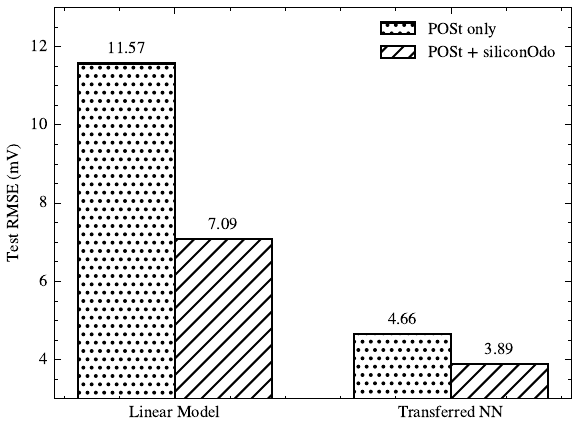}
    \caption{The effectiveness of silicon odometer features.}
    \label{fig:siodo_ablation}
\end{figure}

To further analyze the impact of silicon odometer features, we conducted an ablation study comparing linear regression and our transfer learning model with and without these features, as shown in \cref{fig:siodo_ablation}. The results demonstrate that incorporating silicon odometer data consistently improves prediction accuracy across both linear and non-linear models. This indicates that silicon odometer features provide strong correlations with $\Vmin$ that complement traditional POSt measurements.

These results validate two key aspects of our approach: First, transfer learning effectively leverages knowledge from the 16nm dataset to improve predictions on the 5nm node, particularly valuable when training data is limited. Second, the combination of transfer learning with silicon odometer features provides complementary benefits: the pre-trained model captures general patterns from POSt features, while the silicon odometer data helps adapt to the specific characteristics of the 5nm process. This integration enables robust $\Vmin$ prediction even with minimal training data from the target technology node.

In summary, our experimental findings confirm that advanced neural network models, especially when augmented with silicon odometer data, substantially improve the prediction accuracy of $\Vmin$ on a limited dataset. The results validate the superior efficacy and data efficiency of our approach, which are critical for ensuring the reliable performance of automotive chips in manufacturing processes.